\documentclass[10pt,twocolumn,letterpaper]{article}

\usepackage{wifs}
\usepackage{times}
\usepackage{epsfig}
\usepackage{graphicx}
\usepackage{amsmath}
\usepackage{amssymb}
\usepackage{xcolor,colortbl}

\usepackage[pagebackref=true,breaklinks=true,letterpaper=true,colorlinks,bookmarks=false]{hyperref}

\wifsfinalcopy 

\def\wifsPaperID{10} 

\ifwifsfinal\fi

\definecolor{Gray}{gray}{0.95}

\begin{document}

\title{MesoNet: a Compact Facial Video Forgery Detection Network}

\author{
Darius Afchar\\
\'Ecole des Ponts Paristech\\
Marne-la-Vall\'ee, France\\
{\tt\small darius.afchar@live.fr}
\and
Vincent Nozick\\
JFLI, CNRS, UMI 3527, Japan\\
LIGM, UMR 8049,\\ UPEM, France\\
{\tt\small vincent.nozick@u-pem.fr}
\and
Junichi Yamagishi, Isao Echizen\\
National Institute of Informatics\\
Tokyo, Japan\\
{\tt\small jyamagis@nii.ac.jp,}\\{\tt\small iechizen@nii.ac.jp}
}

\maketitle

\begin{abstract}

This paper presents a method to automatically and efficiently detect face tampering in videos, and particularly focuses on two recent techniques used to generate hyper-realistic forged videos: Deepfake and Face2Face. Traditional image forensics techniques are usually not well suited to videos due to the compression that strongly degrades the data. Thus, this paper follows a deep learning approach and presents two networks, both with a low number of layers to focus on the mesoscopic properties of images. We evaluate those fast networks on both an existing dataset and a dataset we have constituted from online videos. The tests demonstrate a very successful detection rate with more than 98\% for Deepfake and 95\% for Face2Face.

\end{abstract}

\section{Introduction}
Over the last decades, the popularization of smartphones and the growth of social networks have made digital images and videos very common digital objects. According to several reports, almost two billion pictures are uploaded everyday on the internet. This tremendous use of digital images has been followed by a rise of techniques to alter image contents, using editing software like Photoshop for instance. The field of digital image forensics research is dedicated to the detection of image forgeries in order to regulate the circulation of such falsified contents. There have been several approaches to detect image forgeries~\cite{Farid2009,Redi2011}, most of them either analyze inconsistencies relatively to what a normal camera pipeline would be or rely on the extraction of specific image alterations in the resulting image. Among others, image noise~\cite{julliand_IWDW_2015} has been shown to be a good indicator to detect splicing (copy-past from an image to another). The detection of image compression artifacts~\cite{barni2017aligned} also presents some precious hints about image manipulation. 

Today, the danger of \textit{fake news} is widely acknowledged and in a context where more than 100 million hours of video content are watched daily on social networks, the spread of falsified video raises more and more concerns. While significant improvements have been made for image forgery detection, digital video falsification detection still remains a difficult task. Indeed, most methods used with images can not be directly extended to videos, which is mainly due to the strong degradation of the frames after video compression. Current video forensic studies~\cite{milani2012overview} mainly focus on the video re-encoding~\cite{wang2006exposing} and video recapture~\cite{wang2008detecting,lee2010screenshot}, however video edition is still challenging to detect.

For the last years, deep learning methods has been successfully employed for digital image forensics. Amongst others, Barni et al.~\cite{barni2017aligned} use deep learning to locally detect double JPEG compression on images. Rao and Ni~\cite{rao2016deep} propose a network to detect image splicing. Bayar and Stamm~\cite{bayar2016deep} target any image general falsification. Rahmouni et al.~\cite{Rahmouni_WIFS_2017} distinguish computer graphics from photographic images. It clearly appears that deep learning performs very well in digital forensics, and disrupts traditional signal processing approaches. 

In the other hand, deep learning can also be used to falsify videos. Recently, a powerful tool called \textit{Deepfake} has been designed for face capture and reenactment. This methods, initially devoted to the creation of adult content, has not been presented in any academic publication. \textit{Deepfake} follows \textit{Face2Face}~\cite{thies2016face2face}, a non deep learning method introduced by Thies et al. that targets similar goal, using more conventional real-time computer vision techniques. 

This paper addresses the problem of detecting these two video editing processes, and is organized as follows: Sections~\ref{sec:deepfakes} and~\ref{sec:face2face} present more details on \textit{Deepfake} and \textit{Face2Face}, with a special attention for the first one that has not been published. In Section~\ref{sec:proposed_method}, we propose several deep learning networks to successfully overcome these two falsification methods. Section~\ref{sec:results} presents a detailed evaluation of those networks, as well as the datasets we assembled for training and testing. Up to our knowledge, there is no other method dedicated to the detection of the \textit{Deepfake} video falsification technique.


\subsection{Deepfake}\label{sec:deepfakes}

\textit{Deepfake} is a technique which aims to replace the face of a targeted person by the face of someone else in a video. It first appeared in autumn 2017 as a script used to generate face-swapped adult contents. Afterwards, this technique was improved by a small community to notably create a user-friendly application called \textit{FakeApp}.

The core idea lies in the parallel training of two auto-encoders. Their architecture can vary according to the output size, the desired training time, the expected quality and the available resources. Traditionally, an auto-encoder designates the chaining of an encoder network and a decoder network. The purpose of the encoder is to perform a dimension reduction by encoding the data from the input layer into a reduced number of variables. The goal of the decoder is then to use those variables to output an approximation of the original input. The optimization phase is done by comparing the input and its generated approximation and penalizing the difference between the two, typically using a $L^2$ distance. In the case of the \textit{Deepfake} technique, the original auto-encoder is fed with images of resolution $64 \times 64 \times 3 = 12,288$ variables, encodes those images on $1024$ variables and then generates images with the same size as the input.   

\begin{figure}[ht]
\begin{center}
  \includegraphics[width=1.0\linewidth]{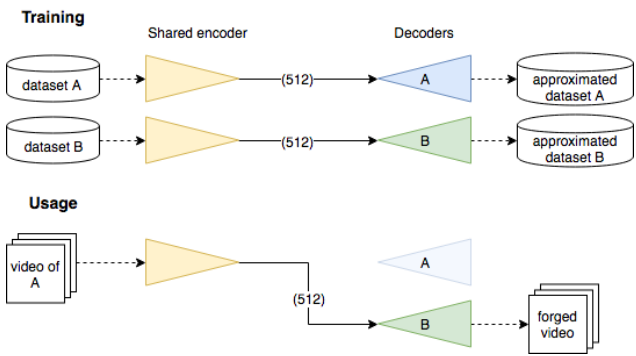}
\end{center}
  \caption{\textit{Deepfake} principle. Top: the training parts with the shared encoder in yellow. Bottom: the usage part where images of~$\mathbf{A}$ are decoded with the decoder of~$\mathbf{B}$.}
\label{fig:deepfake_principle}
\end{figure}

\begin{figure}[ht]
\begin{center}
  \includegraphics[width=1.0\linewidth]{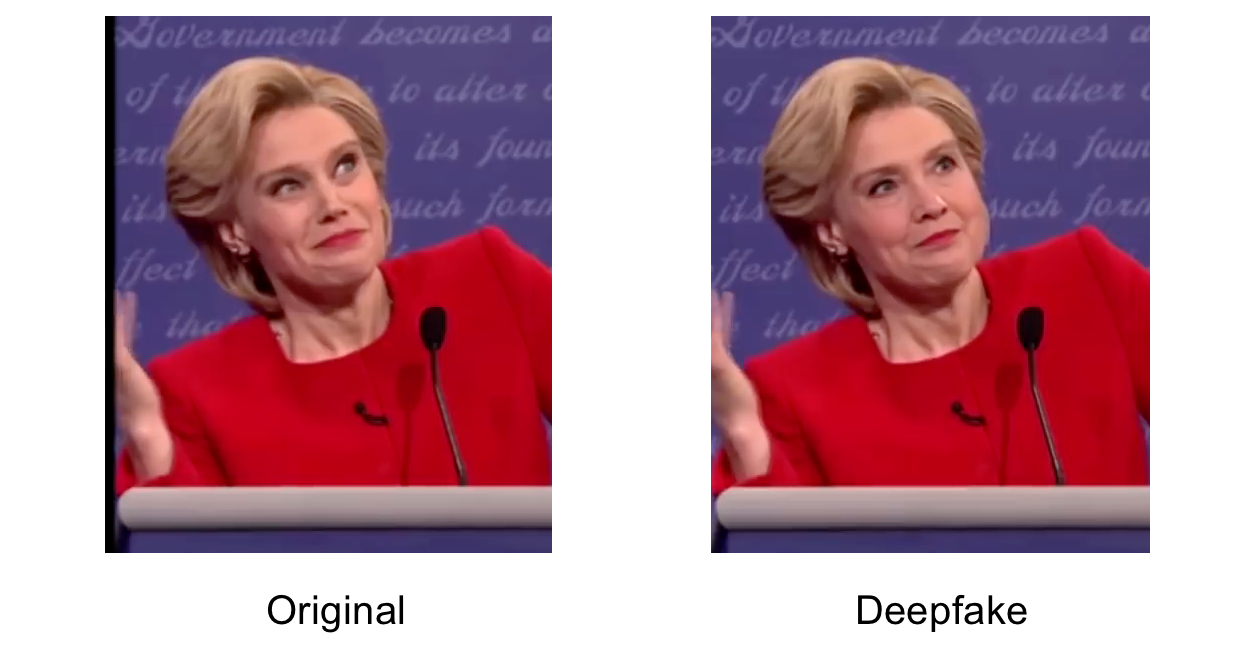}
\end{center}
  \caption{Example of an image (left) being forged (right) using the \textit{Deepfake} technique. Note that the forged face lacks the expressiveness of the original.}
\label{fig:deepfake_principle2}
\end{figure}

The process to generate \textit{Deepfake} images is to gather aligned faces of two different people $\mathbf{A}$ and $\mathbf{B}$, then to train an auto-encoder~$\mathbf{E_A}$ to reconstruct the faces of $\mathbf{A}$ from the dataset of facial images of $\mathbf{A}$, and an auto-encoder $\mathbf{E_B}$ to reconstruct the faces of $\mathbf{B}$ from the dataset of facial images of~$\mathbf{B}$. The trick consists in sharing the weights of the encoding part of the two auto-encoders $\mathbf{E_A}$ and $\mathbf{E_B}$, but keeping their respective decoder separated. Once the optimization is done, any image containing a face of $\mathbf{A}$ can be encoded through this shared encoder but decoded with decoder of $\mathbf{E_B}$. This principle is illustrated in~Figure~\ref{fig:deepfake_principle} and \ref{fig:deepfake_principle2}.

The intuition behind this approach is to have an encoder that privileges to encode general information of illumination, position and expression of the face and a dedicated decoder for each person to reconstitute constant characteristic shapes and details of the person face. This may thus separate the contextual information on one side and the morphological information on the other.

In practice, the results are impressive, which explains the popularity of the technique. The last step is to take the target video, extract and align the target face from each frame, use the modified auto-encoder to generate another face with the same illumination and expression, and then merge it back in the video.

Fortunately, this technique is far from flawless. Basically, the extraction of faces and their reintegration can fail, especially in the case of face occlusions: some frames can end up with no facial reenactment or with a large blurred area or a doubled facial contour. However, those technical errors can easily be avoided with more advanced networks. 
More deeply, and this is true for other applications, auto-encoders tend to poorly reconstruct fine details because of the compression of the input data on a limited encoding space, the result thus often appears a bit blurry. A larger encoding space does not work properly since while the fine details are certainly better approximated, on the other hand, the resulting face loses realism as it tends to resemble the input face, i.e. morphological data are passed to the decoder, which is a undesired effect.


\subsection{Face2Face}\label{sec:face2face}
\begin{figure}[ht]
\begin{center}
  \includegraphics[width=1.0\linewidth]{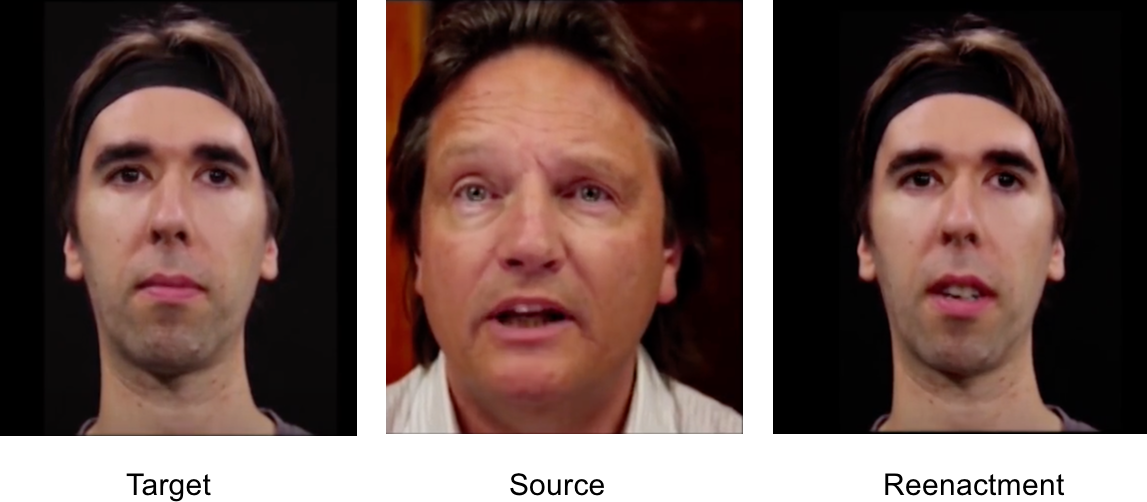}
\end{center}
  \caption{Example of a \textit{Face2Face} reenactment result from the demo video of the Face2Face paper~\cite{thies2016face2face}. 
  }
\label{fig:f2f_principle}
\end{figure}

Reenactment methods, like~\cite{garrido2014automatic}, are designed to transfer image facial expression from a source to a target person. \textit{Face2Face}~\cite{thies2016face2face}, introduced by Thies et al., is its most advanced form. It performs a photorealistic and markerless facial reenactment in real-time from a simple RGB-camera, see~Figure~\ref{fig:f2f_principle}. The program first requires few minutes of prerecorded videos of the target person for a training sequence to reconstruct its facial model. Then, at runtime, the program tracks both the expressions of the source and target actor’s video. The final image synthesis is
rendered by overlaying the target face with a morphed facial blendshape to fit the source facial expression.

\section{Proposed method}\label{sec:proposed_method}

This section presents several effective approaches to deal with either \textit{Deepfake} or \textit{Face2Face}. It turned out that these two problems can not be efficiently solved with a unique network. However, thanks to the similar nature of the falsifications, identical network structures for both problems can yield good results.

We propose to detect forged videos of faces by placing our method at a mesoscopic level of analysis. Indeed, microscopic analyses based on image noise cannot be applied in a compressed video context where the image noise is strongly degraded. Similarly, at a higher — semantic — level, human eye struggles to distinguish forged images~\cite{schetinger2015humans}, especially when the image depicts a human face~\cite{balas2014face,fan2014human}. That is why we propose to adopt an intermediate approach using a deep neural network with a small number of layers.


The two following architectures have achieved the best classification scores among all our tests, with a low level of representation and a surprisingly low number of parameters. They are based on well-performing networks for image classification~\cite{krizhevsky2012imagenet, simonyan2014very} that alternate layers of convolutions and pooling for feature extraction and a dense network for classification. Their source code is available online\footnote{\url{https://github.com/DariusAf/MesoNet}}.

\subsection{Meso-4}

We have started our experiments with rather complex architectures and have gradually simplified them, up to the following one that produces the same results but more efficiently.

This network begins with a sequence of four layers of successive convolutions and pooling, and is followed by a dense network with one hidden layer. To improve generalization, the convolutional layers use ReLU activation functions that introduce non-linearities and Batch Normalization~\cite{ioffe2015batch} to regularize their output and prevent the \textit{vanishing gradient} effect, and the fully-connected layers use Dropout~\cite{srivastava2014dropout} to regularize and improve their robustness.

\begin{figure}[ht]
\begin{center}
      \includegraphics[width=1.0\linewidth]{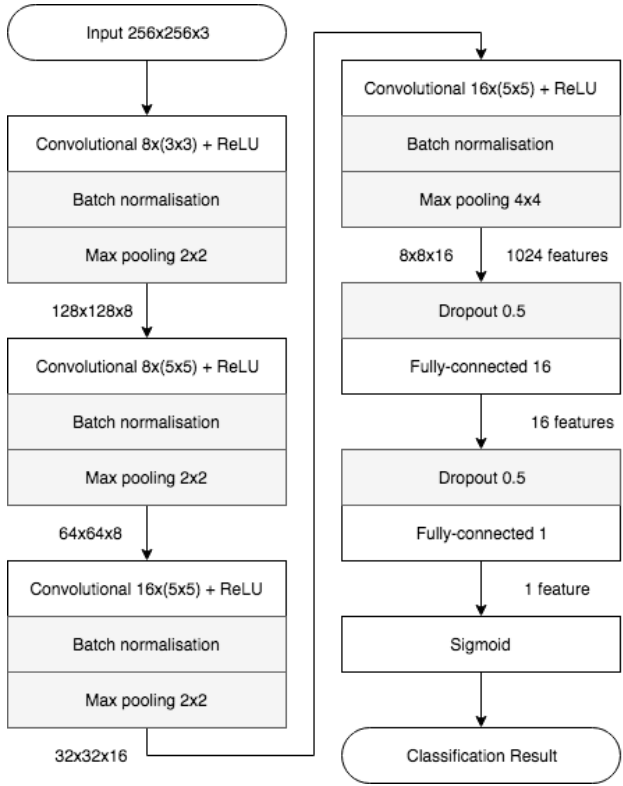}
\end{center}
  \caption{The network architecture of Meso-4. Layers and parameters are displayed in the boxes, output sizes next to the arrows.}
\label{fig:meso4}
\end{figure}

In total, there are 27,977 trainable parameters for this network. Further details can be found on Figure~\ref{fig:meso4}.

\subsection{MesoInception-4}

An alternative structure consists in replacing the first two convolutional layers of Meso4 by a variant of the \textit{inception module} introduced by Szegedy et al~\cite{szegedy2015going}.

\begin{figure}[ht]
\begin{center}
  \includegraphics[width=1.0\linewidth]{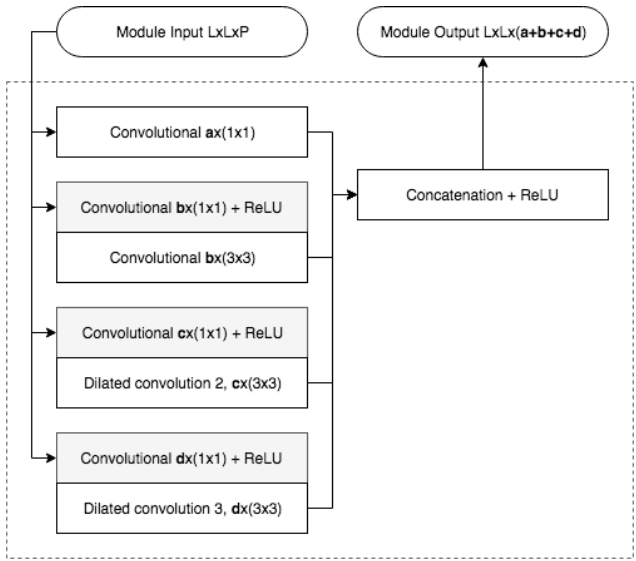}
\end{center}
  \caption{Architecture of the inception modules used in MesoInception-4. The module is parameterized using $a,b,c,d \in \mathbb{N}$. The dilated convolutions are computed without stride.}
\label{fig:mesoinception}
\end{figure}

The idea of the module is to stack the output of several convolutional layers with different kernel shapes and thus increase the function space in which the model is optimized. Instead of the $5\times5$ convolutions of the original module, we propose to use $3\times3$ dilated convolutions~\cite{yu2015multi} in order to avoid high semantic. This idea of using dilated convolutions with the inception module can be found in \cite{shi2017single} as a mean to deal with multi-scale information, but we have added $1\times1$ convolutions before dilated convolutions for dimension reduction and an extra $1\times1$ convolution in parallel that acts as skip-connection between successive modules. Further details can be found in Figure \ref{fig:mesoinception}.

Replacing more than two layers with \textit{inception modules} did not offer better results for the classification. 
The chosen parameters $(a_i,b_i,c_i,d_i)$ for the module at layer $i$ can be found in Table~\ref{table:inception_param}. With those hyper-parameters, this network has 28,615 trainable parameters overall.

\begin{table}[ht]
\begin{center}
    \begin{tabular}{|c|cccc|}
      \hline
      \rowcolor{Gray}
      Layer & $a$ & $b$ & $c$ & $d$  \\
      \hline
      1 & 1 & 4 & 4 & 1 \\
      \hline
      2 & 1 & 4 & 4 & 2 \\
      \hline
    \end{tabular}
\end{center}
\caption{Hyper-parameters for the modified Inception modules used in MesoInception-4}
\label{table:inception_param}
\end{table}

\section{Experiments}\label{sec:results}

In this section we expose the results of the implementation of the two proposed architectures to detect the studied digital forgeries. In order to extend our work to the real case of online videos, we also discuss the robustness of our method to video compression.

\subsection{Datasets}

\subsubsection{Deepfake dataset}

To our knowledge, no dataset gathers videos generated by the \textit{Deepfake} technique, so we have created our own.

Training auto-encoders for the forgery task requires several days of training with conventional processors to achieve realistic results and can only be done for two specific faces at a time. To have a sufficient variety of faces, we have rather chosen to download the profusion of videos available to the general public on the internet.

Thus, 175 rushes of forged videos have been collected from different platforms. Their duration ranges from two seconds to three minutes and have a minimum resolution of $854 \times 480$ pixels. All videos are compressed using the H.264 codec but with different compression levels, which puts us in real conditions of analysis. An accurate study on the effect of compression levels is conducted on another dataset introduced in Section~\ref{sec:face2face_classication}.

All the faces have been extracted using the Viola-Jones detector~\cite{viola2001rapid} and aligned using a trained neural network for facial landmark detection~\cite{dlib09}. In order to balance the distribution of faces, the number of selected frames for extraction per video is proportional to the number of camera angle and illumination changes on the target face. As a reference, approximately 50 faces were extracted per scene.

The dataset has then been doubled with real face images, also extracted from various internet sources and with the same resolutions. Finally, it has been manually reviewed to remove misalignment and wrong face detection. As much as possible, the same distribution of good resolution and poor resolution images were used in both classes to avoid bias in the classification task.

Precise numbers of the image count in each classes as long as the separation into a set used for training and for model evaluation can be found in Table~\ref{table:card_dataset}.

\subsubsection{Face2Face dataset}\label{sec:face2face_classication}

Additionally to the \textit{Deepfake} dataset, we have examined whether the proposed architecture could be used to detect other face forgeries. As a good candidate, the FaceForensics  dataset~\cite{rossler2018faceforensics} contains over a thousand forged videos and their original using the \textit{Face2Face} approach. This dataset is already split into a training, validation and testing set.

More than extending the use of the proposed architecture to another classification task, one advantage of the FaceForensics set is to provide losslessly compressed videos, which has enabled us to evaluate the robustness of our model with different compression levels. To be able to compare our results with those from the FaceForensics paper~\cite{rossler2018faceforensics}, we have chosen the same compression rate with H.264: lossless compression, 23 (light compression), 40 (strong compression).

Only 300 videos were used for training out of more than a thousand. For the model evaluation, the 150 forged video and their original of the testing set were used. Details about the number of extracted face images for each class can be found in Table~\ref{table:card_dataset}.

\begin{table}[ht]
\begin{center}
    \begin{tabular}{|l|c|c|}
      \hline
      \rowcolor{Gray}
      Set & \textbf{forged} class & \textbf{real} class \\
      \hline
      \textit{Deepfake} training & 5111 & 7250 \\
      \textit{Deepfake} testing & 2889 & 4259 \\
      \hline
      \textit{Face2Face} training & 4500 & 4500 \\
      \textit{Face2Face} testing & 3000 & 3000 \\
      \hline
    \end{tabular}
\end{center}
\caption{Cardinality of each class in the studied datasets. Note that for both datasets, 10\% of the training set was used during the optimization for model validation.}\label{table:card_dataset}
\end{table}

\subsection{Classification Setup}

We denote $\mathcal{X}$ the input set and $\mathcal{Y}$ the output set, the random variable pair $(X,Y)$ taking values in $\mathcal{X} \times \mathcal{Y}$, and $f$ the prediction function of the chosen classifier that takes values in $\mathcal{X}$ to the action set $\mathcal{A}$. The chosen classification task is to minimize the error $\mathcal{E}(f) = \mathbb{E}[l(f(X), Y)]$, with $l(a, y) = \frac 1 2 (a - y)^2$.

Both networks have been implemented with Python~3.5 using the Keras~2.1.5 module~\cite{chollet2015keras}. Weights optimization of the network is achieved with successive batches of 75 images of size $256 \times 256 \times 3$ using ADAM~ \cite{kingma2014adam} with default parameters ($\beta_1 = 0.9$ and $\beta_2 = 0.999$). The initial learning rate of $10^{-3}$ is divided by 10 every 1000 iterations down to $10^{-6}$. To improve generalization and robustness, input batches underwent several slight random transformations including zoom, rotation, horizontal flips, brightness and hue changes.

As both network have a relatively small amount of parameters, few hours of optimization on a standard consumer grade computer were enough to obtain good scores.

\subsection{Image classification results}

Classification scores of both trained network are shown in Table~\ref{table:results_df} for the \textit{Deepfake} dataset. Both networks have reached fairly similar score around 90\% considering each frame independently. We do not expect a higher score since the dataset contains some facial images extracted with a very low resolution.  
\begin{table}[ht]
\begin{center}
    \begin{tabular}{|l|c|c|c|}
      \hline
      \rowcolor{Gray}
      Network & \multicolumn{3}{c|}{Deepfake classification score} \\
      \hline
      Class & forged & real & total \\
      \hline
      Meso-4 & 0.882 & 0.901 & 0.891 \\
      MesoInception-4 & 0.934 & 0.900 & 0.917 \\
      \hline
    \end{tabular}
\end{center}
\caption{Classification scores of several networks on the \textit{Deepfake} dataset, considering each frame independently. 
}\label{table:results_df}
\end{table}

Table \ref{table:results_f2f} presents results for the \textit{Face2Face} forgery detection. We observed a notable deterioration of scores at the strong video compression level. The paper that introduces the FaceForensics dataset used in our tests~\cite{rossler2018faceforensics} presents better classification results using the state-of-the-art network for image classification Xception~\cite{chollet2017xception}. However, with the configuration given by the latter paper, we only managed to fine-tune Xception up to obtain a 96.1\% score at the compression level 0 and 93.5\% score at level 23. It is therefore unclear how to interpret the results.

\begin{table}[ht]
\begin{center}
    \begin{tabular}{|l|c|c|c|}
      \hline
      \rowcolor{Gray}
      Network & \multicolumn{3}{c|}{Face2Face classification score} \\
      \hline
      Compression level & 0 & 23 (light) & 40 (strong) \\
      \hline
      Meso-4 & 0.946 & 0.924 & 0.832 \\
      MesoInception-4 & 0.968  & 0.934 & 0.813 \\
      \hline
    \end{tabular}
\end{center}
\caption{Classification scores of several networks on the \textit{FaceForensics} dataset, considering each frame independently.}\label{table:results_f2f}
\end{table}

\begin{figure}[ht]
\begin{center}
  \includegraphics[width=1.0\linewidth]{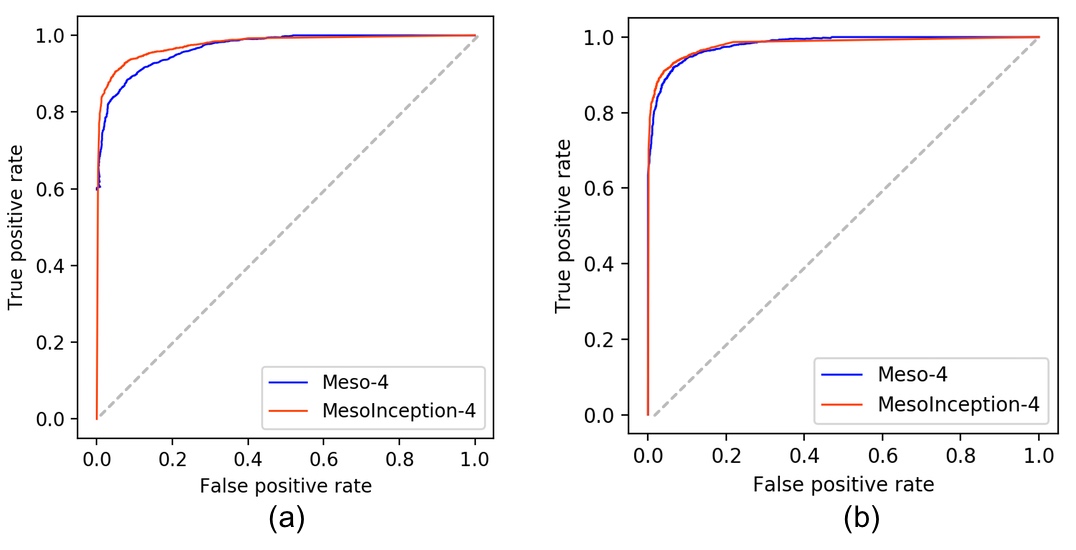}
\end{center}
  \caption{ROC curves of the evaluated classifiers on the \textit{Deepfake} dataset (a) and the \textit{Face2Face} dataset compressed at rate 23 (b).}
\label{fig:ROC_df}
\end{figure}


\subsection{Image aggregation}

One downside of video analysis, especially online videos, is the compression which causes a huge loss of information. But on the other hand, having a succession of frames of the same face makes it possible to multiply experiences and might help obtain a more accurate overall score on the video. A natural way of doing so is to average the network prediction over the video. Theoretically speaking, there is no justification for a gain in scores or a confidence interval indicator as frames of a same video are strongly correlated to one another. In practice, for the viewer comfort, most filmed face contain a majority of stable clear frames. The effect of punctual movement blur, face occlusion and random misprediction can thus be outweighted by a majority of good predictions on a sample of frames taken from the video.

Experimental results can be found in Table~\ref{table:results_video}. The image aggregation significantly improved both detection rate. It even soared higher than \textbf{98\%} with the MesoInception-4 network on the \textit{Deepfake} dataset. Note that on the \textit{Face2Face} dataset, the same score is reached for both networks but the misclassified videos are different.

\begin{table}[ht]
\begin{center}
    \begin{tabular}{|l|c|c|}
      \hline
      \rowcolor{Gray}
      Network & \multicolumn{2}{c|}{Aggregation score} \\
      \hline
      Dataset & Deepfake & Face2Face (23) \\
      \hline
      Meso-4 & \textbf{0.969} & \textbf{0.953} \\
      MesoInception-4 & \textbf{0.984}  & \textbf{0.953} \\
      \hline
    \end{tabular}
\end{center}
\caption{Video classification scores on the two dataset using image aggregation, with the \textit{Face2Face} dataset compressed at rate 23.}\label{table:results_video}
\end{table}

\subsection{Aggregation on intra-frames}

To extend our study of the effect of video compression on forgery detection, we have conducted the same image aggregation but only with intra-frames of compressed videos, i.e. frames that are not interpolated over time, to see if the reduced amount of compression artifacts would help increase the classification score. The flip side is that videos only lasting a few second may contain as little as three I-frames, which cancels out the expected smoothing effect of the aggregation.  

We found out that it rather had a bad effect on the classification, however the different is slight, as shown in Table~\ref{table:ipb_scores}. It might be used as a quick aggregation since the resulting scores are higher than a single image classification. 

\begin{table}[ht]
\begin{center}
    \begin{tabular}{|l|c|c|}
      \hline
      \rowcolor{Gray}
      Network & I-Aggregation score & Difference \\
      \hline
      Meso-4 & 0.932 & -0.037 \\
      MesoInception-4 & 0.959  & -0.025 \\
      \hline
    \end{tabular}
\end{center}
\caption{Classification score variation on the \textit{Deepfake} dataset using only I-frames.}\label{table:ipb_scores}
\end{table}

\subsection{Intuition behind the network}

We have tried to understand how those networks solve the classification problem. This can be done by interpreting weights of the different convolutional kernel and neurons as image descriptors. For instance, a sequence of a positive weight, a negative one, then a positive one again, can be interpreted as a discrete second order derivation. However, this is only relevant for the first layer and does not tell much in the case of faces. 

Another way is to generate an input image that maximizes the activation of a specific filter~\cite{erhan2009visualizing} to see what kind of signal it is reacting to. Concisely, if we note $f_{ij}$ the output of filter $j$ of layer $i$ and $x$ a random image, and we add a regularization on the input to reduce noise, that idea boils down to the maximization of $E(x) = f_{ij}(x) - \lambda \|x\|_p$.

\begin{figure}[ht]
\begin{center}
      \includegraphics[width=1.0\linewidth]{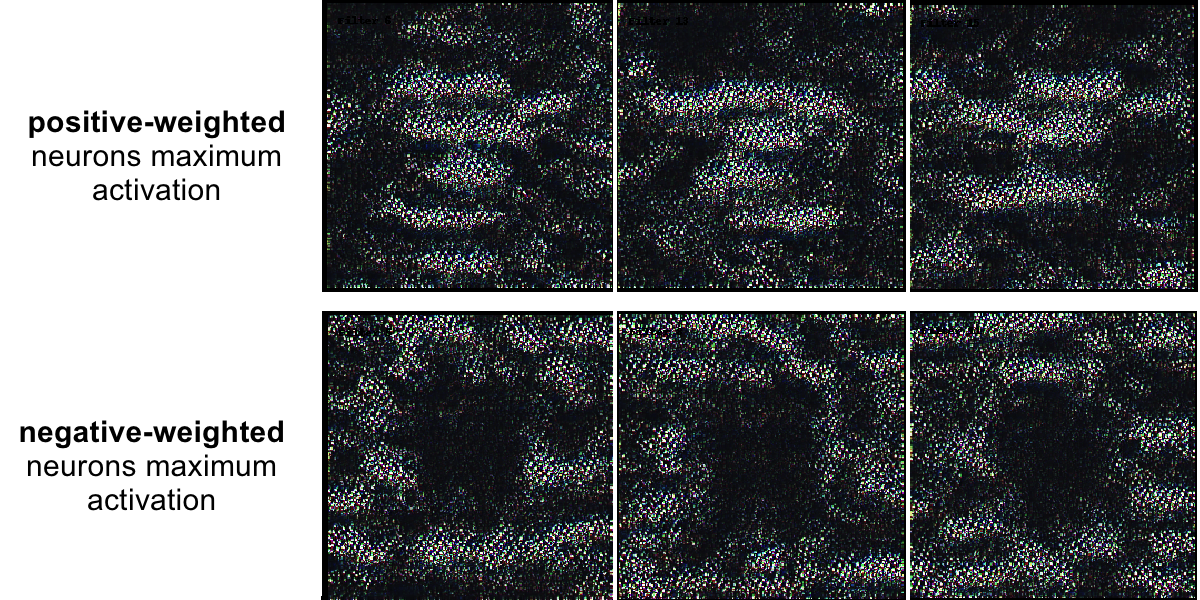}
\end{center}
  \caption{Maximum activation of some neurons of the hidden layer of Meso4. The optimization was done with $\lambda = 10$ and $p = 6$.}
\label{fig:filtersmeso}
\end{figure}

Figure \ref{fig:filtersmeso} shows such maximum activation for several neurons of the last hidden layer of Meso4. We can separate those neurons according to the sign of the weight applied to their output for the final classification decision, thus accounting for whether their activation pushes toward a negative score, which corresponds to the forged class, or a positive one for the real class. Strikingly, positive-weighted neurons activation display images with highly detailed eyes, nose and mouth areas while negative-weighted ones display strong details on the background part, leaving a smooth face area. That's understandable as \textit{Deepfake}-generated faces tend to be blurry, or at least to lack details, compared to the rest of the image that was left untouched.

\begin{figure}[ht]
\begin{center}
      \includegraphics[width=1.0\linewidth]{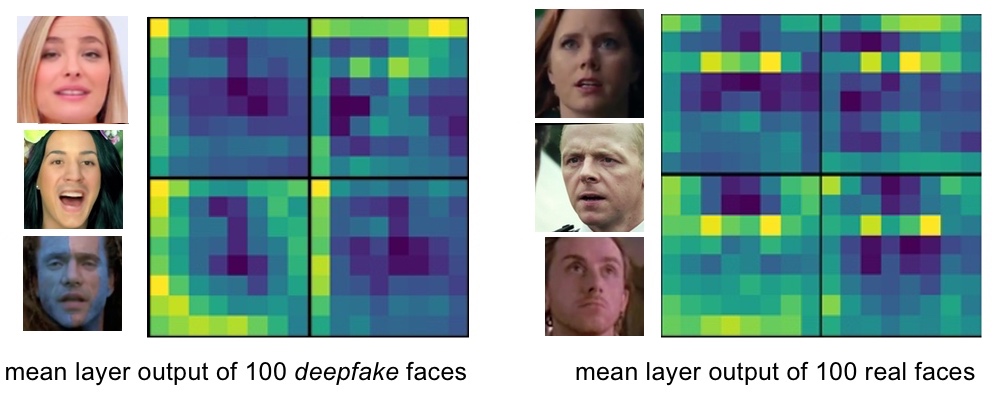}
\end{center}
  \caption{Mean outputs of the \textit{Deepfake} dataset for the some filters of the last convolutional layer of MesoInception-4.}
\label{fig:meanoutput}
\end{figure}

We can also take the mean output of a layer for batches of real and forged images, observe the differences of activation and interpret the parts of the input images that play a key role in the classification. If we study the trained MesoInception-4 network on the \textit{deepfake} dataset, as it can be seen in Figure~\ref{fig:meanoutput}, eyes are strongly activated for real images but not on \textit{deepfake} images for which the background shows the highest peaks. We can surmise that it is again a question of blurriness: the eyes being the most detailed part of real images while it's the background in forged images because of the dimension reduction underwent by the face.

\section{Conclusion}
These days, the dangers of face tampering in video are widely recognized.
We provide two possible network architectures to detect such forgeries efficiently and with a low computational cost. In addition, we give access to a dataset devoted to the \textit{Deepfake} approach, a very popular yet under-documented topic to our knowledge. Our experiments show that our method has an average detection rate of 98\% for \textit{Deepfake} videos and 95\% for \textit{Face2Face} videos under real conditions of diffusion on the internet.


One fundamental aspect of deep learning is to be able to generate a solution to a given problem without the need of a prior theoretical study. However, it is vital to be able to understand the origin of this solution in order to evaluate its qualities and limitations, which is why we spent a significant time visualizing the layers and filters of our networks. We have notably understood that the eyes and mouth play a paramount role in the detection of faces forged with \textit{Deepfake}. We believe that more tools will emerge in the future toward an even better understanding of deep networks to create more effective and efficient ones.

{\small
\bibliographystyle{ieee}
\bibliography{egbib}
}

\end{document}